\documentclass[letterpaper, 10 pt, conference]{ieeeconf}
\IEEEoverridecommandlockouts
\usepackage{multirow}
\usepackage{amsmath,amssymb}
\usepackage{booktabs}
\usepackage{float}
\usepackage{tabularx}
\usepackage{caption}
\usepackage{makecell}
\usepackage{graphicx}
\usepackage{hyperref}
\usepackage{mathptmx}
\usepackage{newtxtext,newtxmath}

\usepackage{caption}
\title{\LARGE \bf
BiView-Touch: Learning Bimanual Tactile Representations by Cross-Hand Completion
}
\author{Chenxin Liang$^{1,2*}$, Youchen Lai$^{1,2*}$, Chuqiao Lyu$^{2}$, Tianxing Chen$^{3,2}$, Shoujie Li$^{4}$, Wenbo Ding$^{1,2}$
\thanks{$^{*}$Contribute equally to this work.}%
\thanks{$^{1}$Tsinghua University.}%
\thanks{$^{2}$XSpark AI.}%
\thanks{$^{3}$The University of Hong Kong.}%
\thanks{$^{4}$Nanyang Technological University.}%
}

\begin{document}
\raggedbottom
\maketitle
\thispagestyle{empty}
\pagestyle{empty}

\begin{abstract}
Bimanual interaction produces complementary tactile views of the same physical process, yet existing tactile representation learning largely models the two hands independently or combines them only for downstream prediction, leaving their cross-hand relationship unexplored. To exploit this overlooked structure, we introduce BiView-Touch, a tactile-only framework that completes masked target-hand latents from the remaining visible target-hand regions and the synchronized full contralateral hand. A student encoder with a geometry-conditioned directional decoder predicts full-view EMA latent targets, while temporal and layout counterfactuals encourage sensitivity to synchronized and anatomically organized source information. Controlled ablations and source-context interventions show that BiView-Touch learns structured cross-hand dependence on temporally aligned and anatomically organized contralateral tactile context, rather than benefiting from bilateral input alone. On the public HumanTouch dataset, its frozen representations consistently outperform representative self-supervised baselines across low-label settings. With only 5\% downstream labels, BiView-Touch achieves relative balanced-accuracy gains of 7.1\% on bilateral wrist-motion recognition and 14.1\% on force-derived interaction-phase recognition. We further introduce BVT-20, a 20-task bilateral tactile dataset, and demonstrate transfer across recording sessions and pretraining corpora, including transfer to a held-out bimanual task. Our code and dataset details are available on the anonymous project page: \url{https://anonymous.4open.science/w/biview-touch-review-site-050C/}.
\end{abstract}

\section{Introduction}

Tactile sensing~\cite{li2026biomimetic,10175024} provides direct observations of physical interaction, including contact, slip, and force cues~\cite{chen2018tactile,dong2017improved}. Existing tactile representation learning explores masked reconstruction, temporal modeling, and interaction dynamics, but largely focuses on individual tactile streams~\cite{cao2023learn,zhang2021dynamic,han2025upvital,xu2025exumi}. Bimanual manipulation instead produces two tactile streams coupled by the same physical process. Prior systems use bilateral tactile sensing for perception and control~\cite{lin2023bi,mao2024efficient,gu2025tactilealoha}, while multimodal frameworks further incorporate vision and action~\cite{liu2026star}. However, cross-hand predictive dependence remains comparatively underexplored as a representation-learning signal.

\begin{figure}[!t]
\centering
\includegraphics[width=\columnwidth]{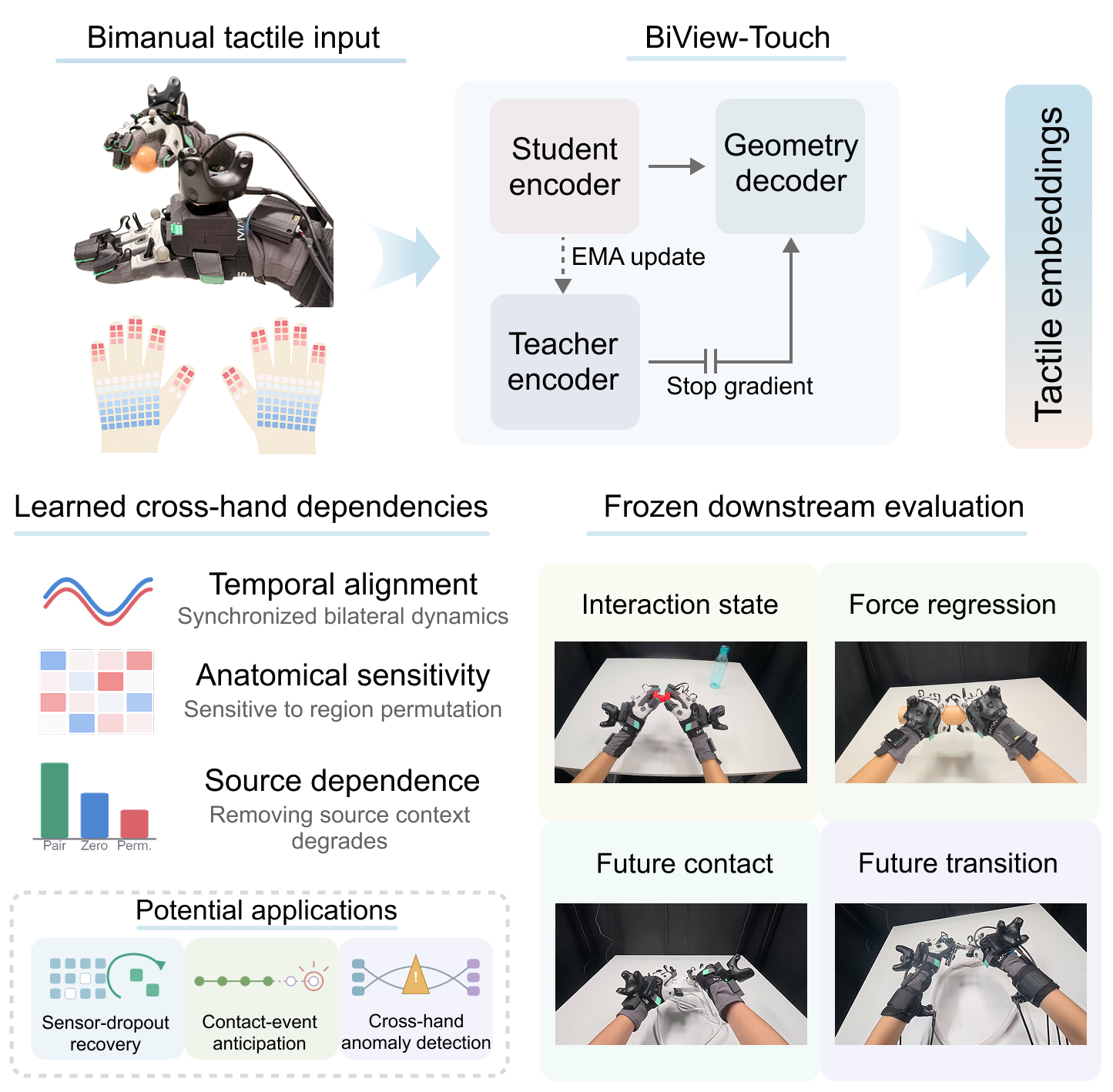}
\caption{Overview of BiView-Touch. Cross-hand latent completion learns region-aware bimanual tactile embeddings that support frozen-backbone recognition, regression, forecasting, and transfer.}
\label{fig}
\end{figure}

To address this problem, we propose \textbf{BiView-Touch}, a tactile-only self-supervised framework that learns bimanual representations through cross-hand completion. During pretraining, a functional region group of one hand is masked throughout a tactile window, while a shared CrossFormer~\cite{zhang2023crossformer} encodes the visible target-hand regions and the complete contralateral hand. Unlike conventional cross-view completion, the masked target-hand content is not directly observed in the contralateral stream; its prediction instead relies on cross-hand dependencies induced by the shared bimanual interaction. A geometry-conditioned cross-attention decoder~\cite{qin2022geometric} predicts the full-view latent representations of the masked regions generated by a continuously updated EMA teacher~\cite{tarvainen2017mean}. As summarized in Fig.~\ref{fig}, the resulting embedding captures temporally aligned, region-selective, and complementary cross-hand information and can be reused
with a frozen backbone for multiple downstream tasks.

\begin{figure*}[t!]
    \centering
    \includegraphics[width=\textwidth]{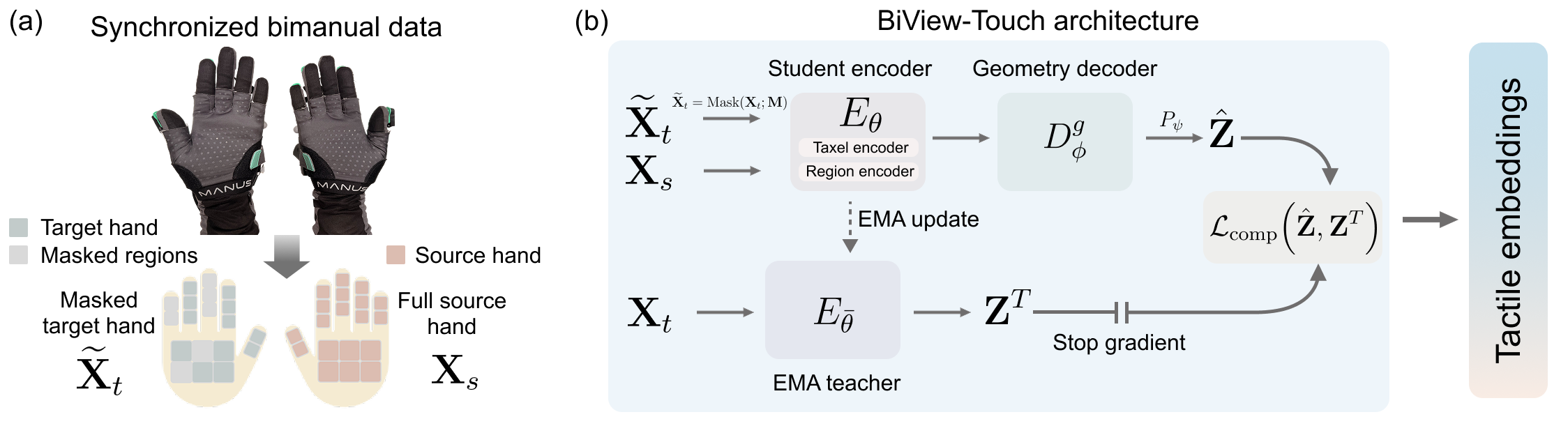}
    \caption{Overview of BiView-Touch. (a) A synchronized bimanual tactile
    window is organized into a masked target hand $\mathbf{X}_t$ and a complete
    contralateral source hand $\mathbf{X}_s$. (b) The shared online encoder and
    geometry-conditioned decoder predict the masked target representation
    $\hat{\mathbf{Z}}$, matched to the stop-gradient full-view target
    $\mathbf{Z}^{T}$ from the EMA encoder. Counterfactual branches reuse the
    online pathway and are omitted for clarity.}
    \label{fig:network}
\end{figure*}

Our contributions are threefold:
\begin{enumerate}

\item We propose \textbf{BiView-Touch}, a self-supervised pretraining framework for learning anatomy-aware bimanual tactile representations through cross-hand completion. Its pretrained encoder-decoder serves as a reusable backbone for downstream bimanual tactile perception.

\item BiView-Touch learns structured cross-hand dependence on temporally aligned and anatomically organized contralateral tactile context. Controlled ablations and source-context interventions verify this dependence, while frozen features achieve relative balanced-accuracy gains of 7.1\% on bilateral wrist-motion recognition and 14.1\% on force-derived interaction-phase recognition with only 5\% downstream labels on HumanTouch.

\item We collect and release BVT-20, a real-world bilateral tactile dataset with 20 bimanual coordination tasks, 44.2 hours of recordings, and 6{,}893 sessions from 22 right-handed participants under paired, role-swapped active--support configurations.

\end{enumerate}

\section{Related Work}

\textbf{Tactile Representation Learning.}
Tactile representation learning has explored supervision from interaction dynamics, masked reconstruction, and temporal prediction. Tactile-glove signals have been shown to capture informative hand--object dynamics beyond direct pressure measurements~\cite{zhang2021dynamic}. Masked reconstruction is another widely used strategy, as in TacMAE~\cite{cao2023learn}, while Luo \textit{et al.}~\cite{luo2024tactile} jointly model task-stage recognition and tactile-dynamics prediction. Sparsh-Skin learns self-supervised representations from distributed tactile-skin signals on dexterous hands~\cite{sharma2025selfsupervised}. AnyTouch further learns unified static--dynamic representations across multiple visuo-tactile sensors~\cite{ICLR2025_4d893f76}. More recent methods also incorporate temporal supervision~\cite{han2025upvital}, canonical geometry and force cues~\cite{wu2025canonical}, sensor-layout priors~\cite{luo2026blind}, or action-conditioned future prediction~\cite{xu2025exumi}. General time-series models provide another baseline family. PatchTST~\cite{Yuqietal-2023-PatchTST} models multivariate sequences through temporal patches. Related representation-learning approaches include self distillation~\cite{assran2023self} and cross-view completion~\cite{weinzaepfel2022croco}, which motivate parts of our latent prediction and decoder design. 

\textbf{Bimanual and Multimodal Tactile Learning.}
Bilateral tactile sensing has been explored in bimanual manipulation frameworks including Bi-Touch~\cite{lin2023bi}, Mao \textit{et al.}~\cite{mao2024efficient}, TactileAloha~\cite{gu2025tactilealoha}, and Tactile Hide and Seek~\cite{fu2025tactile}. Multimodal approaches such as M2VTP~\cite{liu2024masked} and VITaL~\cite{george2025vital} use visuo-tactile pretraining, while 3D-ViTac~\cite{pmlr-v270-huang25e} integrates tactile and visual observations in a unified 3D representation for dexterous bimanual manipulation. VTAO-BiManip~\cite{sun2025vtao} and STAR~\cite{liu2026star} further incorporate vision, action, or object information for bimanual representation learning and manipulation. Recent human-centered datasets such as EgoTouch additionally provide large-scale synchronized bimanual tactile observations for vision-to-touch estimation~\cite{zhou2026touchanything}.

\section{Method}

BiView-Touch learns directional bimanual tactile representations by completing masked target-hand latents from visible target regions and synchronized contralateral context. The two directions, $L\leftarrow R$ and $R\leftarrow L$, share parameters and alternate during training. As shown in Fig.~\ref{fig:network}, an online encoder, interaction decoder, and predictor are trained against full-view latent targets from an EMA encoder~\cite{tarvainen2017mean}.

\begin{figure*}[t]
\centering
\includegraphics[width=0.94\textwidth]{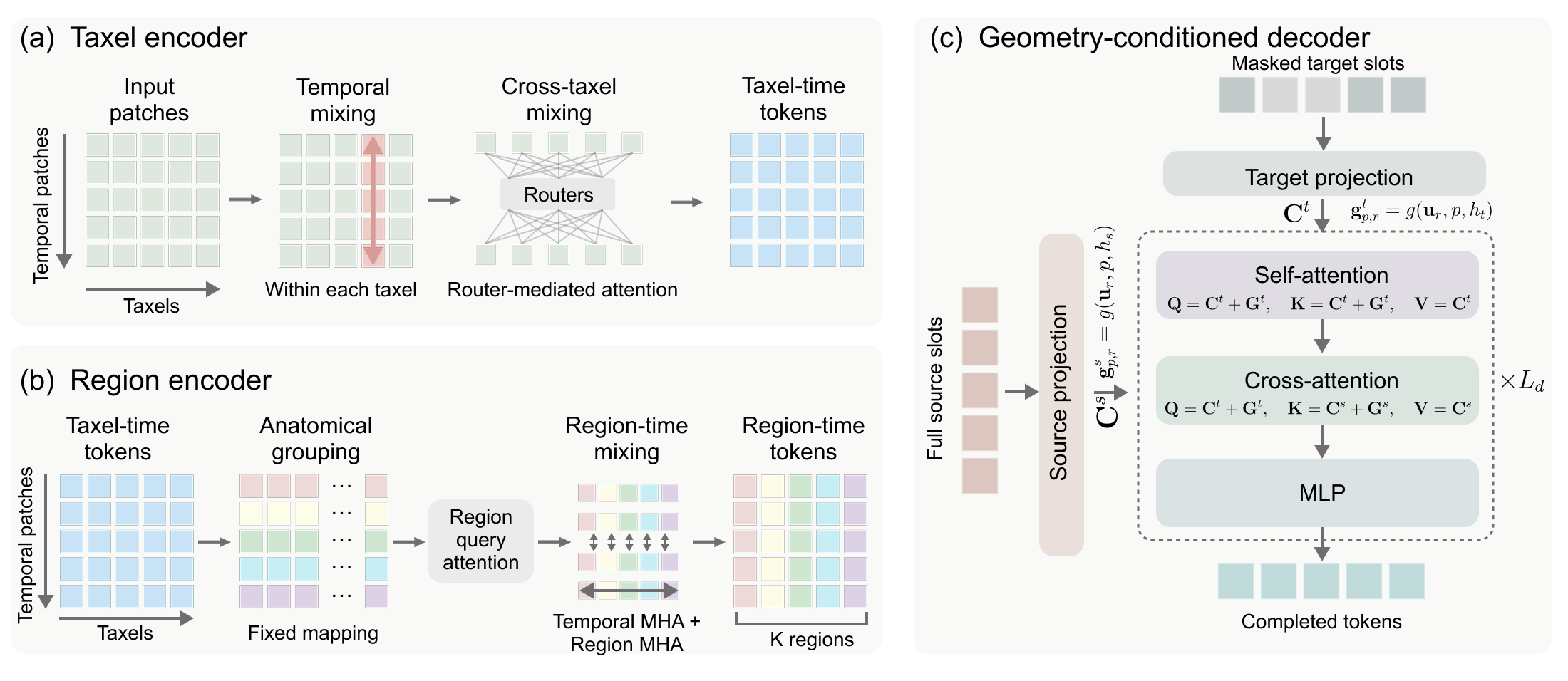}
\caption{BiView-Touch encoder and interaction decoder. (a) CrossFormer-based temporal and cross-taxel encoding produces taxel--time tokens. (b) Anatomy-constrained region queries aggregate taxels into region--time tokens. (c) The geometry-conditioned decoder predicts masked target tokens from target context and contralateral source features using self- and cross-attention.}
\label{fig:encoder_decoder}
\end{figure*}

\subsection{Problem Formulation}

Let $\mathbf{X}^{L},\mathbf{X}^{R}\in\mathbb{R}^{T\times C}$ denote synchronized tactile windows from the two hands, with $T=60$ frames and $C=290$ taxels per hand. Taxels follow the same mirrored canonical order and are grouped into $K=11$ anatomical regions; each window is divided into $P=12$ temporal patches. For target hand $h\in\{L,R\}$ and source hand $\bar h$, a mask $\mathbf{M}\in\{0,1\}^{P\times K}$ specifies hidden target locations. A shared encoder processes the masked target and complete source. The interaction decoder produces completed region--time tokens $\mathbf{R}^{h\leftarrow\bar h}\in\mathbb{R}^{P\times K\times d_d}$, which are mapped by a predictor to
$\hat{\mathbf{Z}}_{h\leftarrow\bar h}\in\mathbb{R}^{P\times K\times d_d}$. No explicit one-to-one correspondence between homologous regions is imposed.

\subsection{Canonical Anatomy-Aware Encoder}

The encoder consists of a taxel encoder followed by an anatomy-aware region encoder. The taxel encoder first divides each taxel sequence into temporal patches using the dimension-segment-wise embedding of CrossFormer~\cite{zhang2023crossformer}. Two two-stage-attention blocks model temporal and cross-taxel dependencies, producing taxel--time tokens
$\mathbf{S}^{h}\in\mathbb{R}^{P\times C\times d_e}$.
Learned taxel, region, and temporal embeddings are added before attention, while masked target patches use a shared learned mask token~\cite{He_2022_CVPR}.

The region encoder aggregates taxel tokens into anatomical regions using learned region queries. For each region $r$, attention is restricted to its taxel set $\mathcal{I}^{h}_{r}$:
\begin{equation}
\mathbf{a}^{h}_{p,r}
=
\operatorname{Attn}\!\left(
\mathbf{q}_{r},
\mathbf{S}^{h}_{p,\mathcal{I}^{h}_{r}},
\mathbf{S}^{h}_{p,\mathcal{I}^{h}_{r}}
\right),
\end{equation}
where $\mathbf{q}_{r}$ is the learned query for region $r$. Factorized temporal and region attention then produces
$\mathbf{U}^{h}\in\mathbb{R}^{P\times K\times d_e}$,
where $\mathbf{U}^{h}_{p,r}$ represents region $r$ at patch $p$.

\subsection{Geometry-Conditioned Interaction Decoder}

Target and source region--time features are projected to the decoder space as
$\mathbf{C}^{t}=\Pi_t(\mathbf{U}^{h})$ and
$\mathbf{C}^{s}=\Pi_s(\mathbf{U}^{\bar h})$, where $\Pi_t$ and $\Pi_s$ are
target- and source-role projections. Masked target locations use a shared
decoder mask token. As shown in Fig.~\ref{fig:encoder_decoder}(c), each decoder
block applies target self-attention, target-to-source cross-attention, and an MLP.

Each region--time location is assigned a geometry embedding
\begin{equation}
\mathbf{g}^{h}_{p,r}
=
\operatorname{LN}\!\left(
\mathbf{e}^{\mathrm{time}}_{p}
+
\mathbf{e}^{\mathrm{hand}}_{h}
+
\operatorname{MLP}\!\left[
\mathbf{u}_{r},\gamma(\mathbf{u}_{r})
\right]
\right),
\end{equation}
where $\operatorname{LN}$ denotes layer normalization,
$\mathbf{e}^{\mathrm{time}}_{p}$ and $\mathbf{e}^{\mathrm{hand}}_{h}$ encode
temporal position and hand identity, $\mathbf{u}_{r}$ is the canonical 2D
centroid of region $r$, and $\gamma(\mathbf{u}_{r})$ is its Fourier
encoding~\cite{NEURIPS2020_55053683}. Stacking these embeddings gives
$\mathbf{G}^{t},\mathbf{G}^{s}\in\mathbb{R}^{P\times K\times d_d}$.
Geometry is added to the query and key streams of cross-attention, while the
value stream contains tactile features only.

The decoder output is
$\mathbf{R}^{h\leftarrow\bar h}
=D_{\phi}^{g}(\mathbf{U}^{h},\mathbf{U}^{\bar h})
\in\mathbb{R}^{P\times K\times d_d}$.
A lightweight predictor $P_{\psi}$ maps it to the unit-normalized latent
prediction
$\hat{\mathbf{Z}}_{h\leftarrow\bar h}
=P_{\psi}(\mathbf{R}^{h\leftarrow\bar h})$.

\subsection{Functional Masking and EMA Targets}

\begin{figure}[H]
\centering
\includegraphics[width=\columnwidth]{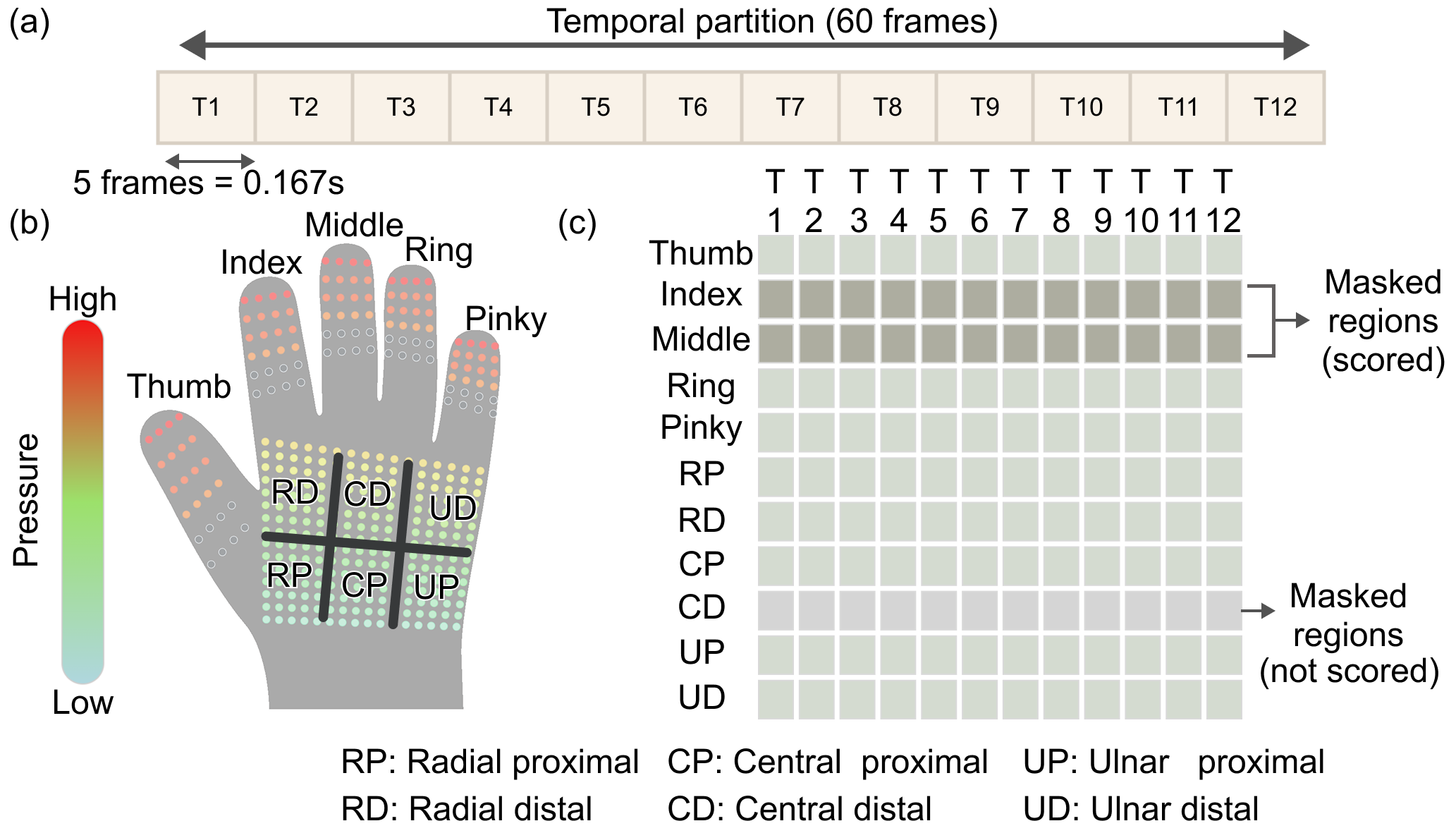}
\caption{Structured functional masking. (a) A 60-frame window is divided into 12 temporal patches. (b) Taxels form 11 anatomical regions. (c) One functional group is masked and scored; additional masked regions act as distractors.}
\label{fig:functional_masking}
\end{figure}

For each target-hand window, one functional region group is masked throughout the temporal window, while additional regions may be masked as distractors. The primary groups include individual fingertips, adjacent fingers, palm regions, proximal--distal palm pairs, and finger--palm combinations. Only the primary masked group contributes to the prediction loss, as illustrated in
Fig.~\ref{fig:functional_masking}.

The target encoder $E_{\bar{\theta}}$ processes the complete, unmasked target hand and is updated as an EMA of the online encoder $E_{\theta}$~\cite{tarvainen2017mean}. For each masked region-time token $i$, the target representation is
\begin{equation}
\mathbf{z}^{T}_{i}
=
\operatorname{sg}\!\left[
\operatorname{norm}_{2}\!\left(
A\!\left[
E_{\bar{\theta}}(\mathbf{X}^{h})_{i}
-
E_{\bar{\theta}}(\mathbf{0})_{i}
\right]
\right)
\right],
\end{equation}
where $\mathbf{0}$ denotes a zero input in the normalized input space, $A$ is a frozen projection head consisting of LayerNorm followed by a linear layer, $\operatorname{norm}_{2}$ denotes $\ell_2$ normalization, and $\operatorname{sg}$ denotes stop-gradient. Subtracting the zero-input response removes input-independent components of the encoder output, yielding a normalized full-view target for latent completion.

\subsection{Counterfactual Correspondence Objective}

For each masked target window, three source contexts are used: the synchronized source $(+)$, a temporally shifted within-session source $(\Delta)$, and a layout counterfactual $(\pi)$ that permutes encoded source-region contents while keeping decoder geometry fixed.

For sample $b$, let $\Omega_b$ denote the set of primary masked region--time
tokens. The prediction error of branch $c\in\{+,\Delta,\pi\}$ is
\begin{equation}
\mathcal{E}^{c}_{b}
=
\frac{1}{|\Omega_b|}
\sum_{i\in\Omega_b}
\left(
2-2(\hat{\mathbf{z}}^{c}_{b,i})^{\top}\mathbf{z}^{T}_{b,i}
\right),
\end{equation}
where $\hat{\mathbf{z}}^{c}_{b,i}$ and $\mathbf{z}^{T}_{b,i}$ are
unit-normalized predicted and target latents, respectively.

The training objective combines synchronized completion with temporal and layout ranking losses:
\begin{equation}
\begin{aligned}
\mathcal{L}_{\mathrm{comp}}
&=
\frac{1}{B}\sum_b \mathcal{E}_b^{+},\\
\mathcal{L}_{\mathrm{sync}}
&=
\frac{1}{B}\sum_b
\left[
\mu_{\mathrm{sync}}
+\mathcal{E}_b^{+}
-\mathcal{E}_b^{\Delta}
\right]_{+},\\
\mathcal{L}_{\mathrm{layout}}
&=
\frac{1}{B}\sum_b
\left[
\mu_{\mathrm{layout}}
+\mathcal{E}_b^{+}
-\mathcal{E}_b^{\pi}
\right]_{+},\\
\mathcal{L}
&=
\mathcal{L}_{\mathrm{comp}}
+
\lambda_{\mathrm{sync}}\mathcal{L}_{\mathrm{sync}}
+
\lambda_{\mathrm{layout}}\mathcal{L}_{\mathrm{layout}},
\end{aligned}
\end{equation}
where $B$ is the batch size and $[x]_{+}=\max(0,x)$. We use
$\mu_{\mathrm{sync}}=0.1$, $\mu_{\mathrm{layout}}=0.05$,
$\lambda_{\mathrm{sync}}=1$, and $\lambda_{\mathrm{layout}}=0.5$ throughout pretraining. The layout-ranking loss is treated as an auxiliary regularizer and is therefore assigned a smaller weight. These objectives favor the synchronized source to produce more accurate completion than temporally mismatched or anatomically permuted alternatives, encouraging sensitivity to temporal alignment and source content-anatomy assignment.

\section{Experiments}

Our experiments test whether BiView-Touch learns transferable and structured cross-hand representations. HumanTouch evaluates low-label frozen features and component contributions, while BVT-20 tests cross-corpus, held-out-task, and future-state transfer. Source interventions further verify dependence on temporally aligned and anatomically organized contralateral context rather than bilateral input alone.

\subsection{Datasets and Protocol}

\subsubsection{HumanTouch}

As a public benchmark for bilateral tactile representation learning, HumanTouch~\cite{humantouch2026} provides approximately 100 hours of recordings and 13{,}469 episodes across 10 contact-rich manipulation tasks. It includes synchronized bilateral whole-hand tactile signals together with hand motion, wrist pose, and visual observations. Only the bilateral tactile recordings are used for pretraining, comprising approximately 89.6 hours of training data and 7.6 hours of validation data under a session-disjoint split. All sessions used for downstream evaluation are excluded from pretraining.

HumanTouch also provides calibrated force estimates in newtons, obtained by calibrating each tactile patch against a reference force sensor. The calibration maps readings from the same tactile array rather than a separate runtime sensor, so force-derived labels are obtained from the same tactile measurements used as model input. For each 60-frame window, let $E$ and $L$ denote the mean $\log(1+\mathrm{force})$ over the first and last 15 frames, with $c=\log(1+0.1)$ and $\delta=0.1$; the six interaction phases follow Table~\ref{tab:phase_rules}, and contact is positive when $L\geq c$. Wrist motion is categorized as stationary, left-only, right-only, or bilateral using MANUS skeletal tracking, which is measured independently of the tactile input. A wrist is active if its displacement over the final 500\,ms exceeds 0.02\,$\mathrm{m}$ or its rotation exceeds 0.15\,$\mathrm{rad}$. Tracking is used only for label generation.

\begin{table}[t]
\caption{Force-derived interaction-phase rules.}
\label{tab:phase_rules}
\centering
\small
\setlength{\tabcolsep}{10pt}
\begin{tabular}{ll}
\toprule
\multicolumn{1}{c}{\textbf{Class}} &
\multicolumn{1}{c}{\textbf{Rule}} \\
\midrule
No contact     & $E<c,\ L<c$ \\
Stable contact & $E\geq c,\ L\geq c,\ |L-E|\leq\delta$ \\
Onset          & $E<c,\ L\geq c$ \\
Loading        & $E\geq c,\ L\geq c,\ L-E>\delta$ \\
Unloading      & $E\geq c,\ L\geq c,\ L-E<-\delta$ \\
Release        & $E\geq c,\ L<c$ \\
\bottomrule
\end{tabular}
\end{table}

\begin{figure}[t]
    \centering
    \includegraphics[width=\columnwidth]{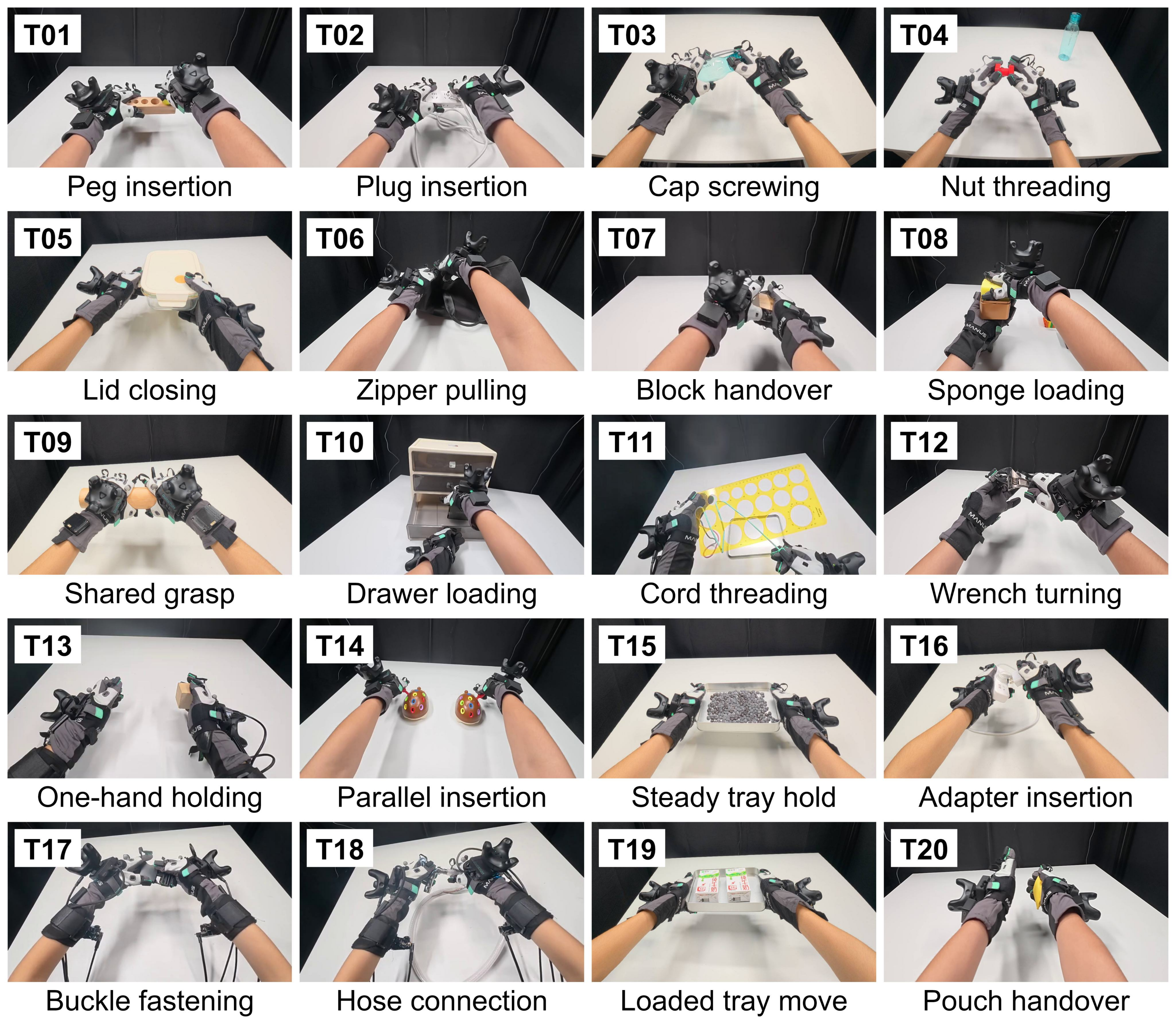}
    \caption{The 20 tasks in our self-collected bimanual tactile dataset,
    spanning insertion and fastening, handover and loading, tool-mediated
    manipulation, shared stabilization, and bimanual transport. Each task is
    recorded under both active--support role configurations.}
    \label{fig:self_collected_tasks}
\end{figure}

\subsubsection{BiView-Touch 20-Task Dataset (BVT-20)}

We collect BVT-20, a bilateral tactile dataset covering 20 bimanual coordination tasks, as illustrated in Fig.~\ref{fig:self_collected_tasks}. Each task is recorded under paired active--support configurations with right-hand active/left-hand support and the reverse. The dataset contains 44.2 hours and 6{,}893 sessions from 22 unique right-handed participants. Task T17 is held out entirely for task-transfer evaluation. Additional acquisition details are provided on the project website.

Because BVT-20 has no calibrated force measurements, interaction labels are derived from tactile signals. Wrist-motion labels used in the context-dependence evaluation are derived independently from MANUS skeletal tracking and are not used as model input. For each hand independently, regional tactile signals and training-split thresholds are used to derive six interaction phases and binary contact. Labels for a target hand are computed solely from that hand and never use the contralateral hand. The resulting labels are uncalibrated tactile proxies and should not be directly compared in absolute accuracy with HumanTouch.

\subsubsection{Training and Evaluation Protocol}

\textbf{Training.} BiView-Touch uses $d_e=192$, $d_d=128$, and four decoder blocks. Models are pretrained for 100 epochs with AdamW using global batch sizes of 128 on HumanTouch and 64 on BVT-20, with a learning rate of $2\times10^{-4}$. One functional group is masked per window, with additional regions masked independently with probability $0.25$. The dense-fusion variant replaces target-to-source cross-attention with an MLP over concatenated target and source region tokens and is pretrained separately under the same protocol. The EMA momentum follows a cosine schedule from 0.996 to 1.0, with AdamW weight decay 0.05.

\textbf{Evaluation.} For downstream evaluation, the pretrained backbone is frozen and only a 193K-parameter readout is optimized on identical session-level label subsets. No temporal or regional average pooling is applied; all region--time decoder representations $\mathbf{R}$ are retained. Table~\ref{tab:readout_protocol} summarizes the readout protocols. Results are averaged over three independently initialized head seeds.

\begin{table}[H]
\centering
\small
\setlength{\tabcolsep}{0pt}
\renewcommand{\arraystretch}{1.08}
\caption{Frozen-feature downstream readout protocols.}
\label{tab:readout_protocol}
\begin{tabular}{lcc}
\toprule
\multicolumn{1}{c}{\textbf{Method}} &
\textbf{Decoder representation} &
\textbf{Readout} \\
\midrule

Within-hand, per-hand
& $\mathbf{R}^{L,\mathrm{wh}},\mathbf{R}^{R,\mathrm{wh}}$
& $H_L,H_R$ \\

Within-hand, bilateral
& $\mathbf{R}^{L,\mathrm{wh}},\mathbf{R}^{R,\mathrm{wh}}$
& $[\mathbf{R}^{L,\mathrm{wh}};\mathbf{R}^{R,\mathrm{wh}}]\!\rightarrow\!H$ \\

\textbf{BiView-Touch}
& $\mathbf{R}^{L\leftarrow R},\mathbf{R}^{R\leftarrow L}$
& $[\mathbf{R}^{L\leftarrow R};\mathbf{R}^{R\leftarrow L}]\!\rightarrow\!H$ \\
\bottomrule
\end{tabular}
\end{table}

Here $\mathbf{U}^{h}$ denotes the encoder output. The within-hand baseline produces $\mathbf{R}^{h,\mathrm{wh}}
=D_{\phi}^{\mathrm{wh}}(\mathbf{U}^{h})$,
where $D_{\phi}^{\mathrm{wh}}$ retains geometry conditioning, self-attention, and MLP blocks but has no contralateral input or cross-attention. BiView-Touch instead produces
$\mathbf{R}^{h\leftarrow\bar h}
=D_{\phi}^{g}(\mathbf{U}^{h},\mathbf{U}^{\bar h})$.
Thus, both methods expose decoder-level frozen representations, while only BiView-Touch performs cross-hand interaction within the pretrained backbone.

\subsection{Representation quality and label efficiency}

\subsubsection{Frozen Representation Quality and Label Efficiency}

All evaluations in this subsection use HumanTouch, on six-way force-derived interaction phase, binary contact, and four-way bilateral wrist motion. Label budgets are sampled at the session level and shared across methods. We compare against a capacity-matched raw MLP, an end-to-end dual-branch CNN~\cite{luo2024tactile}, PatchTST-SSL~\cite{Yuqietal-2023-PatchTST}, adapted Sparsh-Skin~\cite{sharma2025selfsupervised}, and a within-hand EMA baseline using the same encoder family without contralateral pretraining, evaluated with both per-hand and bilateral-concatenation readouts. We separately pretrain dense-fusion and cross-attention variants of BiView-Touch under the same masking and optimization protocol.

\begin{table*}[t]
\caption{Low-label recognition on HumanTouch (test balanced accuracy, \%, mean over three runs). $\dagger$: frozen backbone with a 193K-parameter head; $\ddagger$: supervised training from random initialization; $\ast$: dense-fusion architecture pretrained as a separate checkpoint.}
\label{tab:label_efficiency}
\centering
\small
\setlength{\tabcolsep}{3.5pt}
\renewcommand{\arraystretch}{1.20}

\begin{tabular*}{\textwidth}{
@{\extracolsep{\fill}}llccccccccc@{}
}
\toprule
\multicolumn{2}{c}{\multirow{2}{*}{{\raisebox{-0.6ex}{\textbf{Method}}}}}
& \multicolumn{3}{c}{\textbf{Force-derived phase (6-way)\, $\uparrow$}}
& \multicolumn{3}{c}{\textbf{Contact (binary)\, $\uparrow$}}
& \multicolumn{3}{c}{\textbf{Wrist motion (4-way)\, $\uparrow$}} \\
\cmidrule(lr){3-5}
\cmidrule(lr){6-8}
\cmidrule(lr){9-11}
&
& \textbf{5\%} & \textbf{10\%} & \textbf{25\%}
& \textbf{5\%} & \textbf{10\%} & \textbf{25\%}
& \textbf{5\%} & \textbf{10\%} & \textbf{25\%} \\
\midrule

\multicolumn{2}{@{}l}{
Raw tactile + MLP$^{\ddagger}$ (193K)}
& 39.47 & 42.30 & 46.00
& 71.34 & 73.42 & 75.44
& 51.99 & 56.38 & 61.51 \\

\multicolumn{2}{@{}l}{
Raw Dual CNN$^{\ddagger}$ (1.60M)~\cite{luo2024tactile}}
& 43.82 & 46.95 & 49.62
& 74.25 & 75.71 & 77.25
& 62.67 & 64.13 & 66.64 \\

\midrule

\multicolumn{2}{@{}l}{
PatchTST-SSL$^{\dagger}$~\cite{Yuqietal-2023-PatchTST}}
& 33.29 & 35.76 & 39.76
& 66.41 & 68.28 & 70.72
& 51.67 & 55.33 & 60.20 \\

\multicolumn{2}{@{}l}{
Sparsh-Skin adapted$^{\dagger}$~\cite{sharma2025selfsupervised}}
& 38.47 & 42.46 & 46.42
& 71.16 & 73.78 & 75.94
& 60.76 & 62.08 & 65.27 \\

\multirow{2}{*}{Within-hand EMA$^{\dagger}$}
& Per-hand head
& 36.56 & 39.43 & 43.13
& 67.38 & 69.77 & 72.14
& 60.95 & 63.12 & 65.21 \\

& Bilateral concat
& 35.43 & 39.28 & 43.22
& 67.17 & 69.86 & 72.38
& 59.73 & 63.03 & 66.24 \\

\addlinespace[2pt]

\multirow{2}{*}{\textbf{BiView-Touch}$^{\dagger}$}
& \textbf{Dense fusion$^{\ast}$}
& \textbf{44.27} & \textbf{47.15} & \textbf{50.68}
& \textbf{73.04} & \textbf{75.24} & \textbf{77.42}
& \textbf{65.40} & \textbf{67.78} & \textbf{70.70} \\

& \textbf{Cross-attention}
& \textbf{43.88} & \textbf{46.88} & \textbf{50.46}
& \textbf{72.81} & \textbf{74.92} & \textbf{77.12}
& \textbf{65.27} & \textbf{68.37} & \textbf{71.13} \\

\bottomrule
\end{tabular*}
\end{table*}

Table~\ref{tab:label_efficiency} shows that bilateral input alone does not explain the gains. PatchTST-SSL remains weaker despite receiving both hands during pretraining, while adding bilateral concatenation only at downstream readout changes the within-hand EMA baseline by at most $\pm1.2$~pp and does not close the gap to BiView-Touch. The two BiView-Touch fusion variants perform similarly overall, indicating that the main benefit comes from cross-hand pretraining rather than a specific fusion operator. At 5\% labels, the cross-attention variant achieves relative bAcc gains of 7.1\% on wrist motion and 14.1\% on phase recognition over representative frozen SSL baselines, while training only the downstream head.

\subsubsection{Pretraining Component Ablations}

In order to validate our design, we evaluate the three components introduced specifically for cross-hand pretraining. The Full model in this ablation study is an independently rerun pretraining instance with a different pretraining seed from Table~\ref{tab:label_efficiency}. All comparisons below are therefore made only within the matched ablation group.

\textbf{Synchronization loss.} Table~\ref{tab:sync_ablation} shows that removing $\mathcal{L}_{\mathrm{sync}}$ consistently reduces label efficiency, particularly for interaction-phase recognition. We further apply a whole-source temporal-shift intervention, in which the contralateral tactile stream is temporally misaligned while the target input and downstream head remain fixed. Under this intervention, the model trained with $\mathcal{L}_{\mathrm{sync}}$ drops by 4.0--6.1~pp, whereas the model trained without $\mathcal{L}_{\mathrm{sync}}$ is nearly invariant, showing that the synchronization loss promotes global temporal pairing across hands.

\begin{table}[H]
\centering
\small
\setlength{\tabcolsep}{5.5pt}
\renewcommand{\arraystretch}{1.0}
\caption{Ablation of the synchronization objective on HumanTouch frozen-feature recognition (balanced accuracy, \%).}
\label{tab:sync_ablation}

\begin{tabular}{>{\raggedright\arraybackslash}p{2.15cm}cccc}
\toprule
\multirow{2}{=}{\centering\textbf{Method}}
& \multicolumn{2}{c}{\textbf{Phase bAcc} $\uparrow$}
& \multicolumn{2}{c}{\textbf{Wrist motion bAcc} $\uparrow$} \\
\cmidrule(lr){2-3}\cmidrule(lr){4-5}
& 5\% & 25\% & 5\% & 25\% \\
\midrule
w / o $\mathcal{L}_{\mathrm{sync}}$
& 37.13 & 44.19 & 60.16 & 66.86 \\
\textbf{w / $\mathcal{L}_{\mathrm{sync}}$}
& \textbf{44.12} & \textbf{49.91} & \textbf{63.30} & \textbf{69.01} \\
\bottomrule
\end{tabular}
\end{table}

\textbf{Functional-group masking.}
An ablation study is conducted to assess the contribution of functional-group masking relative to random masking. Regional alignment gain $G$ is defined as the bAcc difference between correctly aligned contralateral context and a 120-frame shift of the same source region, and $\Delta G$ measures the difference in this gain between the two masking strategies. Table~\ref{tab:mask_ablation} shows positive phase $\Delta G$ at the 25\% and 100\% label budgets, while contact effects are mixed. An exploratory stratification into dynamic phases (onset, loading, unloading, release) and static phases (no-contact, stable-contact) further shows a positive interaction at 25\%, $\Delta G_{\mathrm{dynamic}}-\Delta G_{\mathrm{static}}=0.337$~pp (95\% CI [0.114, 0.565]), indicating stronger temporal alignment for evolving interaction states.

\begin{table}[t]
\centering
\small
\setlength{\tabcolsep}{7.5pt}
\renewcommand{\arraystretch}{1.12}
\caption{Effect of functional-group masking on regional temporal alignment.
$\Delta G$ denotes the difference in regional alignment gain between functional-group and random masking.
The last column reports the interaction between dynamic and static phase groups.
Values are percentage points with paired session-level bootstrap 95\% confidence intervals.}
\label{tab:mask_ablation}

\begin{tabular}{cccc}
\toprule
\multicolumn{1}{c}{\textbf{Budget}} &
\multicolumn{1}{c}{\makecell{\textbf{Phase}\\$\Delta G\,\uparrow$}} &
\multicolumn{1}{c}{\makecell{\textbf{Contact}\\$\Delta G\,\uparrow$}} &
\multicolumn{1}{c}{\makecell{\textbf{Dynamic--static}\\\textbf{interaction} $\uparrow$}} \\
\midrule

5\%
& \makecell{+0.033\\[-1pt]{\scriptsize [-0.063, 0.129]}}
& \makecell{-0.129\\[-1pt]{\scriptsize [-0.223, -0.036]}}
& \makecell{-0.066\\[-1pt]{\scriptsize [-0.296, 0.158]}} \\

25\%
& \makecell{\textbf{+0.290}\\[-1pt]{\scriptsize \textbf{[0.171, 0.414]}}}
& \makecell{+0.107\\[-1pt]{\scriptsize [0.009, 0.202]}}
& \makecell{\textbf{+0.337}\\[-1pt]{\scriptsize \textbf{[0.114, 0.565]}}} \\

100\%
& \makecell{\textbf{+0.203}\\[-1pt]{\scriptsize \textbf{[0.087, 0.320]}}}
& \makecell{+0.045\\[-1pt]{\scriptsize [-0.037, 0.128]}}
& \makecell{+0.166\\[-1pt]{\scriptsize [-0.079, 0.413]}} \\

\bottomrule
\end{tabular}
\end{table}

\textbf{Layout-ranking loss.} We evaluate all 55 pairwise region swaps and 8 global permutations by comparing models trained with and without $\mathcal{L}_{\mathrm{layout}}$. The model trained with $\mathcal{L}_{\mathrm{layout}}$ is markedly more sensitive to fingertip--palm reassignment and global permutations, whereas within-fingertip and within-palm swaps produce only small changes. Across the 55 pairwise swaps at the 25\% label budget, degradation correlates with anatomical distance for the model trained with $\mathcal{L}_{\mathrm{layout}}$ ($\rho=0.454$, $p=0.0072$), but not for the model trained without it ($\rho=-0.111$, $p=0.419$). Jointly permuting source content and geometry changes bAcc by less than 0.04~pp for either model, supporting sensitivity to content-location assignment.

\begin{table}[t]
\centering
\small
\setlength{\tabcolsep}{1.5pt}
\renewcommand{\arraystretch}{1.10}
\caption{Layout-loss ablation under source-region reassignment on HumanTouch. Values are mean bAcc drops (pp); larger values indicate greater sensitivity to content--anatomy mismatch. Adjacent denotes neighboring-region swaps; F$\leftrightarrow$F and P$\leftrightarrow$P denote within-fingertip and within-palm swaps; F$\leftrightarrow$P denotes fingertip--palm swaps; Global averages eight full-region permutations.}
\label{tab:layout_ablation}
\begin{tabular}{ccccccc}
\toprule
\multicolumn{1}{c}{\textbf{Budget}} &
\multicolumn{1}{c}{\textbf{Method}} &
\textbf{Adjacent}$\uparrow$ &
\textbf{F$\leftrightarrow$F}$\uparrow$ &
\textbf{P$\leftrightarrow$P}$\uparrow$ &
\textbf{F$\leftrightarrow$P}$\uparrow$ &
\textbf{Global}$\uparrow$ \\
\midrule
\multirow{2}{*}{5\%}
& w / o $\mathcal{L}_{\mathrm{layout}}$ & 0.008 & 0.004 & 0.022 & 0.075 & 0.232 \\
& \textbf{w /  $\mathcal{L}_{\mathrm{layout}}$} & 0.139 & 0.025 & 0.072 & \textbf{0.632} & \textbf{2.268} \\
\midrule
\multirow{2}{*}{25\%}
& w / o $\mathcal{L}_{\mathrm{layout}}$ & -0.006 & 0.009 & -0.015 & -0.031 & 0.051 \\
& \textbf{w /  $\mathcal{L}_{\mathrm{layout}}$} & 0.229 & 0.004 & 0.071 & \textbf{1.015} & \textbf{2.792} \\
\bottomrule
\end{tabular}
\end{table}

\subsubsection{Complementary Downstream Evaluations}

We further examine whether the frozen representations retain information beyond the force-derived classification targets. BiView-Touch shows the clearest force-regression improvement under masked-region evaluation, reducing MAE and RMSE to 0.892 and 2.388~N, respectively, while remaining comparable to Sparsh-Skin under masked-hand evaluation (Table~\ref{tab:force_tab}). On 26 manually annotated operation stages across four HumanTouch tasks, both BiView-Touch architectures also outperform PatchTST-SSL and the within-hand EMA baseline and remain comparable to adapted Sparsh-Skin and the supervised dual-branch CNN as in Table~\ref{tab:operation_stage}.

\begin{table}[t]
\caption{Frozen-feature force regression on HumanTouch
(N; lower is better).}
\label{tab:force_tab}
\centering
\small
\setlength{\tabcolsep}{0.5pt}
\renewcommand{\arraystretch}{1.15}

\begin{tabular*}{\columnwidth}{
@{\extracolsep{\fill}}lcccc@{}
}
\toprule
\multicolumn{1}{c}{\multirow{2}{*}{\raisebox{-0.6ex}{\textbf{Method}}}}
& \multicolumn{2}{c}{\textbf{Masked region}}
& \multicolumn{2}{c}{\textbf{Masked hand}} \\
\cmidrule(lr){2-3}
\cmidrule(lr){4-5}
& MAE $\downarrow$
& RMSE $\downarrow$
& MAE $\downarrow$
& RMSE $\downarrow$ \\
\midrule

PatchTST-SSL~\cite{Yuqietal-2023-PatchTST}
& 1.026 & 2.845 & 1.115 & 3.175 \\

Sparsh-Skin adapted~\cite{sharma2025selfsupervised}
& 0.937 & 2.558 & 1.080 & 3.075 \\

\textbf{BiView-Touch}
& \textbf{0.892}
& \textbf{2.388}
& \textbf{1.075}
& \textbf{3.065} \\

\bottomrule
\end{tabular*}
\end{table}

\begin{table}[t]
\caption{HumanTouch operation-stage recognition
(\%, mean over four runs). $\dagger$: frozen backbone;
$\ddagger$: end-to-end training; $\ast$: separately pretrained
dense-fusion architecture.}
\label{tab:operation_stage}
\centering
\small
\setlength{\tabcolsep}{1.5pt}
\renewcommand{\arraystretch}{1.15}

\begin{tabular*}{\columnwidth}{
@{\extracolsep{\fill}}llccc@{}
}
\toprule
\multicolumn{2}{c}{\textbf{Method}}
& \textbf{Accuracy}\,$\uparrow$
& \textbf{bAcc}\,$\uparrow$
& \textbf{Macro-F1}\,$\uparrow$ \\
\midrule

\multicolumn{2}{@{}l}{
Raw Dual CNN$^{\ddagger}$~\cite{luo2024tactile}}
& 72.66 & 71.03 & 69.89 \\

\midrule

\multicolumn{2}{@{}l}{
PatchTST-SSL$^{\dagger}$~\cite{Yuqietal-2023-PatchTST}}
& 64.47 & 61.81 & 62.01 \\

\multicolumn{2}{@{}l}{
Sparsh-Skin adapted$^{\dagger}$~\cite{sharma2025selfsupervised}}
& 72.20 & 70.72 & 69.83 \\

\multicolumn{2}{@{}l}{
Within-hand EMA$^{\dagger}$}
& 59.51 & 61.67 & 57.99 \\

\addlinespace[2pt]

\multirow{2}{*}{\textbf{BiView-Touch}$^{\dagger}$}
& \textbf{Dense fusion$^{\ast}$}
& \textbf{72.86}
& \textbf{71.50}
& \textbf{70.29} \\

& \textbf{Cross-attention}
& \textbf{71.71}
& \textbf{71.13}
& \textbf{69.07} \\

\bottomrule
\end{tabular*}
\end{table}

\begin{table*}[t]
\caption{Cross-corpus transfer on BVT-20 (balanced accuracy, \%). Both models use frozen BiView-Touch backbones and complete bilateral input. Best results are shown in bold. }
\label{tab:bvt20_generalization}
\centering
\small
\setlength{\tabcolsep}{3pt}
\renewcommand{\arraystretch}{1.15}

\begin{tabular*}{\textwidth}{
@{\extracolsep{\fill}}llcccccc@{}
}
\toprule
\multicolumn{1}{c}{\multirow{2}{*}{{\raisebox{-0.6ex}{\textbf{Target}}}}}
& \multicolumn{1}{c}{\multirow{2}{*}{\raisebox{-0.6ex}{\textbf{Pretraining dataset}}}}
& \multicolumn{3}{c}{\textbf{Session-disjoint test}}
& \multicolumn{3}{c}{\textbf{Held-out task}} \\
\cmidrule(lr){3-5}
\cmidrule(lr){6-8}
&
& \textbf{5\%}\,$\uparrow$ & \textbf{10\%}\,$\uparrow$ & \textbf{25\%}\,$\uparrow$
& \textbf{5\%}\,$\uparrow$ & \textbf{10\%}\,$\uparrow$ & \textbf{25\%}\,$\uparrow$ \\
\midrule

\multirow{2}{*}{Tactile-derived phase}
& HumanTouch~\cite{humantouch2026}
& $63.49$
& $69.07$
& $73.78$
& $51.54$
& $58.28$
& $63.30$ \\

& BVT-20
& $\mathbf{66.80}$
& $\mathbf{72.24}$
& $\mathbf{77.44}$
& $\mathbf{54.87}$
& $\mathbf{60.76}$
& $\mathbf{66.46}$ \\

\midrule

\multirow{2}{*}{Contact}
& HumanTouch~\cite{humantouch2026}
& $92.21$
& $93.55$
& $94.54$
& $90.02$
& $90.49$
& $92.13$ \\

& BVT-20
& $\mathbf{93.25}$
& $\mathbf{94.34}$
& $\mathbf{95.31}$
& $\mathbf{90.43}$
& $\mathbf{91.36}$
& $\mathbf{93.42}$ \\

\bottomrule
\end{tabular*}
\end{table*}

\subsection{Transfer on BVT-20}
We evaluate BVT-20 transfer across sessions, pretraining corpora, and a held-out task. All experiments freeze the backbone and train the same unpooled head on identical session-level label subsets, averaging over three head seeds. We compare HumanTouch- and BVT-20-pretrained backbones, with downstream heads trained only on BVT-20 labels.

\subsubsection{Cross-corpus Transfer}

Table~\ref{tab:bvt20_generalization} compares the two pretrained backbones on six-way tactile-derived phase recognition and binary contact classification. Although both representations support label-efficient recognition on the session-disjoint BVT-20 test set, in-domain pretraining consistently produces the stronger results. The advantage is more pronounced for tactile-derived phase than for contact, suggesting that basic contact information transfers readily between corpora, whereas fine-grained interaction structure benefits more from domain-matched pretraining. Cross-corpus transfer nevertheless remains effective without adapting the HumanTouch-pretrained backbone. 

\subsubsection{Held-Out-Task Transfer}

Table~\ref{tab:bvt20_generalization} evaluates transfer to a single held-out task, T17, which is excluded entirely from both pretraining and downstream-head training. Both backbones remain above chance, while tactile-derived phase shows a larger generalization gap than contact. At the 25\% label budget, bAcc drops from 77.44\% to 66.46\% for phase, compared with 95.31\% to 93.42\% for contact.

Future-state prediction further probes whether the representation captures interaction dynamics on the held-out task. Table~\ref{tab:future_transition} shows AUPRC above transition prevalence at all horizons, reaching 54.67\% versus 16.18\% at 1000\,ms. Source removal and no-cross-attention reduce performance, but these inference-time interventions are treated as auxiliary evidence because they introduce input or pathway distribution shifts.

\begin{table}[H]
\caption{Future contact-transition prediction on held-out BVT-20 task T17 (\%, mean over three runs). Rate denotes transition prevalence, and AUPRC is reported for the paired context model. Source ablations are inference-time interventions and may introduce distribution shift.}
\label{tab:future_transition}
\centering
\small
\setlength{\tabcolsep}{0.5pt}
\renewcommand{\arraystretch}{1.15}

\begin{tabular*}{\columnwidth}{
@{\extracolsep{\fill}}lccccc@{}
}
\toprule
\multirow{2}{*}{\raisebox{-2.5ex}{\textbf{Horizon}}}
& \multirow{2}{*}{\raisebox{-2.5ex}{\textbf{Rate}}}
& \multicolumn{3}{c}{\textbf{Balanced accuracy}\,$\uparrow$}
& \multirow{2}{*}{\raisebox{-2.5ex}{\textbf{AUPRC}\,$\uparrow$}} \\
\cmidrule(lr){3-5}
&
& \makecell[c]{\textbf{Paired}\\\textbf{context}}
& \makecell[c]{Zero\\source}
& \makecell[c]{No cross-\\attention}
& \\
\midrule

100\,ms
& 3.16
& \textbf{66.21}
& 61.39
& 52.87
& 15.23 \\

250\,ms
& 6.97
& \textbf{78.02}
& 67.40
& 51.73
& 30.71 \\

500\,ms
& 10.73
& \textbf{80.14}
& 71.47
& 52.11
& 42.22 \\

1000\,ms
& 16.18
& \textbf{80.71}
& 73.57
& 52.83
& 54.67 \\

\bottomrule
\end{tabular*}
\end{table}

\subsection{Dependence on Contralateral Context}
\label{sec:context}

To determine whether BiView-Touch learns structured cross-hand dependence rather than merely benefiting from bilateral input, inference-time interventions selectively disrupt the contralateral context while keeping the target input and downstream head fixed. Cross-session substitution, temporal shifting, and region permutation respectively perturb paired interaction context, temporal alignment, and anatomical assignment. Zero-source and no-cross-attention serve as auxiliary destructive controls.

\subsubsection{Cross-Hand Dependence on HumanTouch}

We first evaluate source-context dependence on HumanTouch using a fixed downstream head with the target hand fully masked. Wrist-motion recognition is particularly informative because its labels are derived from MANUS tracking rather than tactile input. As shown in Table~\ref{tab:humantouch_context}, replacing the paired source with a different-stage but motion-matched donor leaves wrist bAcc nearly unchanged, whereas a motion-mismatched donor, source-region permutation, or source removal causes substantial degradation. This indicates that the frozen representation uses task-relevant contralateral motion content rather than source presence alone.

\subsubsection{Context Dependence on BVT-20}

\begin{figure}[H]
    \centering
    \includegraphics[width=0.8\columnwidth]{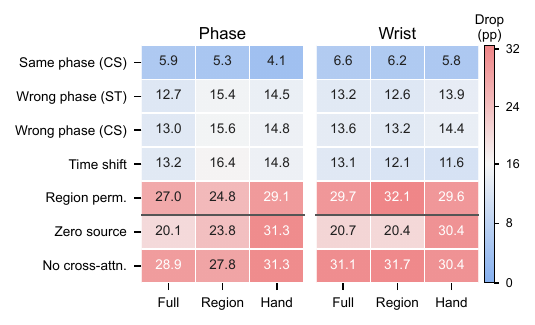}
    \caption{Source-context sensitivity on BVT-20 for phase and wrist-motion recognition. Cells show balanced-accuracy drops relative to correctly paired context under three target-visibility levels. Region permutation disrupts source anatomical assignment, while zero-source and no-cross-attention are auxiliary destructive controls. Full, Region, and Hand denote full target visibility, masked target-region input, and fully masked target-hand input; CS and ST denote cross-session and same-task substitution.}
    \label{fig:bvt_intervention}
\end{figure}

Fig.~\ref{fig:bvt_intervention} shows the same interventions on BVT-20 under three levels of target visibility. Same-phase cross-session substitution causes the smallest degradation, while phase mismatch and temporal misalignment produce larger drops. Region permutation is particularly disruptive for both phase and wrist motion and causes larger drops than source removal when the target remains fully or partially visible. The wrist-motion result is especially informative because its labels are derived independently from MANUS tracking, supporting sensitivity to the anatomical assignment of contralateral tactile content rather than source presence alone.

\begin{table}[t]
\centering
\small
\setlength{\tabcolsep}{20pt}
\renewcommand{\arraystretch}{1.05}
\caption{HumanTouch wrist-motion recognition under source-context interventions with the target hand fully masked. Values are bAcc (\%, mean over three head seeds).}
\label{tab:humantouch_context}
\begin{tabular}{lc}
\toprule
\multicolumn{1}{c}{\textbf{Source context}} & \textbf{Wrist bAcc} $\uparrow$ \\
\midrule
Correct paired & \textbf{54.93} \\
Different stage, same motion & 54.16 \\
Different stage, different motion & 19.31 \\
Region permutation & 31.35 \\
Zero source & 25.00 \\
\bottomrule
\end{tabular}
\end{table}

\section{Conclusions and Limitations}

We introduce BiView-Touch, a tactile-only self-supervised framework for learning bimanual representations through cross-hand latent completion. Frozen BiView-Touch features improve low-label recognition over temporal and within-hand baselines, while source-context interventions show sensitivity to synchronized and anatomically organized contralateral tactile information. However, several limitations remain. Reported variations reflect downstream head seeds rather than independent pretraining runs, and the current BVT-20 splits do not fully disentangle participant, acquisition, and task effects. On HumanTouch, phase and contact labels are derived from calibrated force estimates computed from the same tactile measurements and are therefore neither mutually independent nor independent of the model input; BVT-20 uses uncalibrated tactile-derived proxies. Future work will investigate stronger participant- and task-disjoint evaluation, broader tactile morphologies, and closed-loop robotic manipulation.

\appendix
\subsubsection{BVT-20 Task Definitions}
\label{app:bvt20_tasks}

BVT-20 contains 20 manipulation tasks recorded with synchronized bilateral tactile streams. The tasks span insertion, fastening, handover, supported loading, tool use, stabilization, and transport. Several include controlled misalignment, obstruction, load changes, or motion-intensity changes. Task descriptions organize the dataset and are not used as annotations during pretraining. Table~\ref{tab:bvt20_task_list} summarizes the manipulation sequence of each task.

\begin{table}[!t]
\caption{BVT-20 task definitions. Initial and final idle states are omitted for brevity.}
\label{tab:bvt20_task_list}
\centering
\small
\setlength{\tabcolsep}{2pt}
\renewcommand{\arraystretch}{1.05}
\begin{tabular}{@{}p{0.09\columnwidth}p{0.86\columnwidth}@{}}
\toprule
\makebox[0.09\columnwidth][c]{\textbf{ID}}
& \makebox[0.86\columnwidth][c]{\textbf{Manipulation sequence}} \\
\midrule
T01 & Peg insertion: large- and small-angle failed attempts, followed by aligned insertion and removal. \\
T02 & Three-pin plug insertion: two angular-offset attempts, followed by aligned insertion and unplugging. \\
T03 & Bottle-cap threading: misaligned attempt, back-off, aligned tightening, and removal. \\
T04 & Nut and bolt: align and thread the nut onto the bolt, then unscrew it. \\
T05 & Food-container lid: misaligned closing attempt, realignment, latching, and reopening. \\
T06 & Fabric-bag zipper: open and close the zipper while supporting the bag. \\
T07 & Wooden block: twelve alternating hand-to-hand transfers. \\
T08 & Supported box: insert and remove rigid, sponge, and foam blocks while the other hand holds the box aloft. \\
T09 & Wooden bar: bilateral lifting, grip adjustment, stabilization, and placement. \\
T10 & Drawer loading: two obstructed placements at different angles, followed by aligned placement and retrieval. \\
T11 & Cords and perforated board: thread and retrieve cords while exchanging hand roles. \\
T12 & Wrench and fastener: align the wrench, tighten the fastener, then loosen it. \\
T13 & Asymmetric motion: manipulate a wooden block in one hand while the empty hand makes similar movements. \\
T14 & Dual inserts: synchronously insert and withdraw two inserts from separate objects. \\
T15 & Loaded tray: bilateral support under steady, slight-sway, and pronounced-sway conditions. \\
T16 & Two-pin charger: hand over, plug in, unplug, and return the charger. \\
T17 & Buckle: handovers, a misaligned connection attempt, aligned fastening, and separation. \\
T18 & Hose fitting: misaligned connection attempt, withdrawal, aligned connection, and removal. \\
T19 & Tray transport: move the tray when empty and after successive carton load changes. \\
T20 & Soft pouch: twelve alternating hand-to-hand transfers. \\
\bottomrule
\end{tabular}
\end{table}

Together, these tasks cover alignment and correction, rotational and tool-mediated manipulation, handover, shared support, and asymmetric or synchronized motion. This variety allows cross-hand dependence to be examined across multiple manipulation families rather than a single task type.

\subsubsection{Region-Selective Dependence Across BVT-20 Tasks}
\label{app:region_dependence}

We mask each target region and remove each source region in turn, measuring the increase in latent completion error without a downstream head. The resulting $11\times11$ matrices are normalized by source-region taxel count, then adjusted by removing additive target/source effects and the component jointly explained by three random-initialization controls. Signed entries represent deviations from this fitted baseline, not absolute degradation.

\begin{figure}[H]
\centering
\includegraphics[width=0.9\columnwidth]{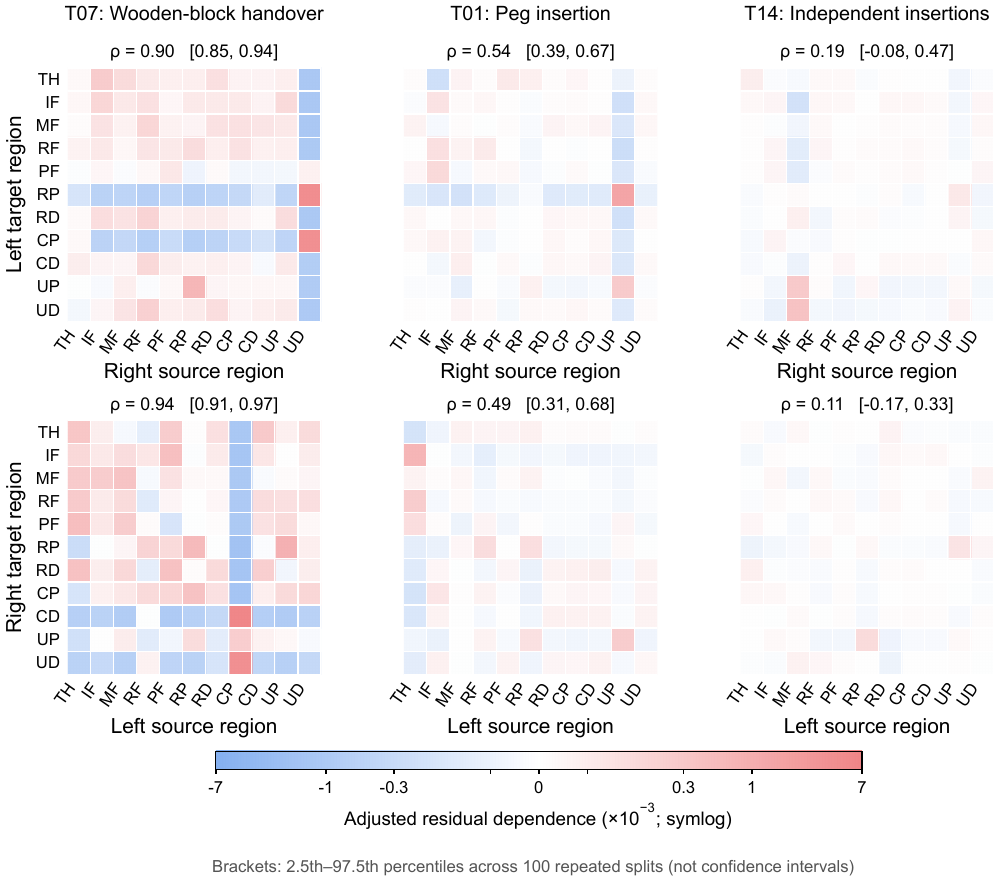}
\caption{Residual region-dependency maps for three BVT-20 tasks after removing additive effects and random-control predictions. Rows/columns denote target/source regions; upper/lower panels predict the left/right hand. Colors share a symmetric-log scale. Labels show mean Spearman $\rho$ and the 2.5th--97.5th percentile range over 100 repeated splits.}
\label{fig:bvt20_region_dependence}
\end{figure}

Figure~\ref{fig:bvt20_region_dependence} presents three examples selected to span observed reproducibility: wooden-block handover (T07), peg insertion (T01), and independent bilateral insertions (T14). Maps use all available sessions. Reproducibility is evaluated over 100 repeated pairs of disjoint 15-session subsets, with adjustments fitted separately within each subset. Mean Spearman correlations are 0.90/0.94, 0.54/0.49, and 0.19/0.11, respectively, for right-to-left/left-to-right prediction.

These descriptive examples illustrate region-selective dependence and its variable stability, not functional correspondence or significant task-category differences. Motion repetition and acquisition conditions remain possible influences. Percentile ranges summarize repeated-split variability, not confidence intervals; low reproducibility does not imply absent contralateral information.

\clearpage
\bibliographystyle{IEEEtran}
\bibliography{ref}
\end{document}